# A Comparison of Nature Inspired Algorithms for Multi-threshold Image Segmentation

**Valentín Osuna-Enciso[1], Erik Cuevas[2], Humberto Sossa[1]**

[1]Centro de Investigación en Computación-IPN
Av. Juan de Dios Batiz S/N
Col. Nueva Industrial Vallejo, México, D. F. MEXICO
hsossa@cic.ipn.mx

[2]Departamento de Ciencias Computacionales
Universidad de Guadalajara, CUCEI
Av. Revolución 1500, Guadalajara, Jal, México
{erik.cuevas, valentin.osuna}@cucei.udg.mx

**Abstract**

In the field of image analysis, segmentation is one of the most important preprocessing steps. One way to achieve segmentation is by mean of threshold selection, where each pixel that belongs to a determined class is labeled according to the selected threshold, giving as a result pixel groups that share visual characteristics in the image. Several methods have been proposed in order to solve threshold selection problems; in this work, it is used the method based on the mixture of Gaussian functions to approximate the 1D histogram of a gray level image and whose parameters are calculated using three nature inspired algorithms (Particle Swarm Optimization, Artificial Bee Colony Optimization and Differential Evolution). Each Gaussian function approximates the histogram, representing a pixel class and therefore a threshold point. Experimental results are shown, comparing in quantitative and qualitative fashion as well as the main advantages and drawbacks of each algorithm, applied to multi-threshold problem.

## 1. Introduction

Nature has been a great source of inspiration for creating metaheuristic algorithms, as can be seen of important proposals in such areas since first evolutive programs were created almost three decades ago [30], leaving clear that even today, this trend is still valid, by the development and use of concepts such as artificial neural networks, evolutionary algorithms, swarming algorithms and so on, not to mention new developments in the computational paradigms mentioned. Particularly, three of those algorithms are Differential Evolution (DE), Particle Swarm Optimization (PSO) and Artificial Bee Colony Optimization (ABC) that have been used to solve difficult optimization problems. DE, originally proposed by Storn and Price in 1995 [28], is a population-based algorithm in which the population is evolved from one generation to the next using special defined operators such as mutation, crossover, and selection. The PSO algorithm, introduced in 1995 [5], is based on swarm behavior of birds and fish where the solutions, called particles, 'fly' through the search space using simple mapping equations; this algorithm has been used to solve distinct optimization problems, being the main vision and video processing [8]. More recently in 2005, the ABC algorithm has been introduced by Karaboga [6]. Such algorithm, inspired by the intelligent behavior of honey-bees, consists of three essential components: food source positions, nectar-amount and honey-bee classes. Each food source position represents a feasible solution for the problem under consideration. The nectar-amount for a food source represents the quality of such solution (represented by fitness value). Each bee-class symbolizes one particular operation for generating new candidate food source positions.The aforementioned algorithms have been used to deal with several optimization problems in the area of image analysis, giving good results in terms of performance [1, 2, 6,31, 22].

Image segmentation has been the subject of intensive research and a wide variety of segmentation techniques has been reported in the last two decades. In general terms, image segmentation divides an image into related sections or regions, consisting of image pixels having related data feature values. It is an essential issue since it is the first step for image understanding and any other, such as feature extraction and recognition, heavily depends on its results. Segmentation algorithms are based on two significant criteria: the homogeneity of a region (thresholding) and the discontinuity between adjacent disjoint regions (finding edges). Since the segmented image obtained from the homogeneity criterion has the advantage of smaller storage space, fast processing speed and ease in manipulation, thresholding techniques are considered the most popular [33].

Thresholding techniques can be classified into two categories: bi-level and multi-level. In the former, one limit value is chosen to segment an image into two classes: one representing the object and the other one segmenting the background. When distinct objects are depicted within a given scene, multiple threshold values have to be selected for proper segmentation, which is commonly called multi-level thresholding. A variety of thresholding approaches have been proposed for image segmentation, including conventional methods [14, 16, 18, 19] and intelligent techniques [20, 22, 1, 2,4,30]. Extending the segmentation algorithms to a multilevel approach may cause some inconveniences: (i) they may have no systematic or analytic solution when the number of classes to be detected increases and (ii) they may also show a slow convergence and/or high computational cost [32].

In this work, the segmentation approach is based on a parametric model composed by a group of Gaussian functions (Gaussian mixture). Gaussian mixture (GM) represents a flexible method of statistical modelling with a wide variety of scientific applications [34, 35]. In general, GM involves the model selection, i.e., to determine the number of components in the mixture (also called model order), and the estimation of the parameters of each component in the mixture that better adjust the statistical model. Computing the parameters of Gaussian mixtures is considered a difficult optimization task, sensitive to the initialization [37] and full of possible singularities [36]. As an optimization problem, the presented here requires an objective function, which makes use of Hellinger distance to compare the GM candidate and the original histogram. This distance measure works with probability density functions, making it appropriate to the problem presented in this work, and was shown that this distance is the most suitable to construct a minimum distance estimator [7]. The Hellinger distance has been used in on-line recognition of handwritten text [21], in signal modulation [29] and classification and localization of underwater acoustic signals [3], only to mention some uses.

This paper presents the use of evolutionary algorithms to compute threshold selection for image segmentation. In this approach, the segmentation process is considered as an optimization problem approximating the 1-D histogram of a given image by means of a Gaussian mixture model whose parameters are calculated through the DE, the PSO and the ABC algorithm. In the model, each Gaussian function approximating the histogram represents a pixel class and therefore a threshold point in the segmentation scheme. Those algorithms are experimentally compared by solving the multi-threshold problem, obtaining in such a way the main advantages and drawbacks of each one.

Previous studies performed to assess the performance of DE, PSO and ABC algorithms included the work in [38] showing that ABC performs better than PSO, and DE on a suite of classical benchmark functions. It was shown that the performance of ABC is better or at least similar than DE and PSO while having a smaller number of parameters to tune. In [11] several DE variants were empirically compared over a benchmark of 13 functions, finding that the version *best/1/bin* has the best behavior regardless quality and robustness. A study comparing variations of PSO over power systems is made in [13], finding that theenhanced general passive congregation PSO shown the best performance, but also has a high computational cost. The performance of DE, PSO and real valued Genetic Algorithm over a benchmark of functions was made in [15], and the best results were obtained in general by DE.The aforementioned studies suffer from one limitation: the comparisons are based on a set of synthetic functions with exact and well-known solutions and none of them were applied to image processing. The proposed study overcomes such drawbacks by assessing the performance of the set of evolutionary algorithms when they are applied to the image processing problem of segmentation, particularly multi-threshold segmentation (the GM estimation), where an exact solution does not exist. The comparison is carried out based on two different statistics namely: the solution reached and the histogram approximation according to a quality measure based on Hausdorff distance among ground-truth images and segmentation results. The versions of algorithms studied in this work are DE (best/1/bin), PSO (attractive/repulsive PSO) and normal ABC.

The remainder of this work is organized as follows: in section 2 we present the method following Gaussian approximation of the histogram, whereas in sections 3, 4 and 5 we show a brief overview of Differential Evolution, Particle Swarm Optimization and Artificial Bee Colony optimization, respectively, as well as some of their implementation details. Experimental results are shown up in Section 6, followed byconclusions in Section 7.

## 2. Gaussian Approximation

In what follows histogram $h(g)$ represents a gray level distribution of an image with $L$ gray levels$[0,1, \ldots, L-1]$; it is also assumed that $h(g)$ is normalized, considered as a probability distribution function:

$$h(g) = \frac{n_g}{N}, \ h(g) \geq 0, \tag{1}$$

$$N = \sum_{g=0}^{L-1} n_g, \ \text{and} \ \sum_{g=0}^{L-1} h(g) = 1,$$

where$n_g$ denotes the number of pixels with gray level $g$,whereas$N$ represents the total number of pixels contained in the image. The mix of Gaussian probability functions:

$$p(x) = \sum_{i=1}^{K} P_i \cdot p_i(x) = \sum_{i=1}^{K} \frac{P_i}{\sqrt{2\pi}\sigma_i} \exp\left[\frac{-(x-\mu_i)^2}{2\sigma_i^2}\right] \tag{2}$$

can approximate the original image histogram, dealing with$P_i$as the a priori probability of class $i$, $p_i(x)$as the probability distribution function of gray-level random variable $x$ in class $i$, $\mu_i$and $\sigma_i$as the mean and standard deviation of the $i$-th probability distribution function and $K$ as the number of classes contained in the image. In addition, the constraint$\sum_{i=1}^{K} P_i = 1$ must be made certain.

The Hellinger distance is used to estimate the 3K ($P_i, \mu_i$and$\sigma_i, i = 1, \ldots, K$) parameters, comparing in such way the mixture of Gaussian functions (or candidate histogram) and the original histogram:

$$E = \sqrt{\sum_{j=1}^{n} \left[\sqrt{p(x_j)} - \sqrt{h(x_j)}\right]^2} \tag{3}$$

where $p(x_j)$ is the histogram formed with the candidate Gaussian mixture and $h(x_j)$ is the experimental histogram that corresponds to the gray level image. Such a formula represents the fitness function used by the three nature inspired algorithms reported in this work and it does not need extra parameters.

The next step is to determine the optimal threshold values. Considering that the data classes are organized such that $\mu_1 < \mu_2 < \cdots < \mu_K$, the threshold values can thus be calculated by estimating the overall probability error for two adjacent Gaussian functions, as follows:

$$E(T_i) = P_{i+1} \cdot E_1(T_i) + P_i \cdot E_2(T_i), \tag{4}$$

$$i = 1, 2, \ldots, K - 1$$

considering

$$E_1(T_i) = \int_{-\infty}^{T_i} p_{i+1}(x) dx, \tag{5}$$

and

$$E_2(T_i) = \int_{T_i}^{\infty} p_i(x) dx, \tag{6}$$

$E_1(T_i)$ is the probability of mistakenly classifying the pixels in the ($i$+1)-th class to the $i$-th class, while $E_2(T_i)$ is the probability of erroneously classifying the pixels in the $i$-th class to the ($i$+1)-th class; as was abovementioned, formulae (4), (5) and (6) states for two consecutive Gaussian functions in order to get one threshold point and the process is repeated for each pair until get all the threshold points. The $P_j$'s are the a-priori probabilities within the combined probability density function, and $T_i$ is the threshold value between the $i$-th and the ($i$+1)-th classes. One $T_i$ value is chosen such as the error $E(T_i)$ is minimized. By differentiating $E(T_i)$ with respect to $T_i$ and equating the result to zero, it is possible to use the following equation to define the optimum threshold value $T_i$:

$$AT_i^2 + BT_i + C = 0 \tag{7}$$

considering

$$\begin{aligned} A &= \sigma_i^2 - \sigma_{i+1}^2 \\ B &= 2 \cdot (\mu_i \sigma_{i+1}^2 - \mu_{i+1} \sigma_i^2) \\ C &= (\sigma_i \mu_{i+1})^2 - (\sigma_{i+1} \mu_i)^2 + 2 \cdot (\sigma_i \sigma_{i+1})^2 \cdot \ln\left(\frac{\sigma_{i+1} P_i}{\sigma_i P_{i+1}}\right) \end{aligned} \tag{8}$$

Even though the above quadratic equation has two possible solutions, only one of them is feasible (the positive one which falls within the interval). Figure 1 shows the determination process of the threshold points.

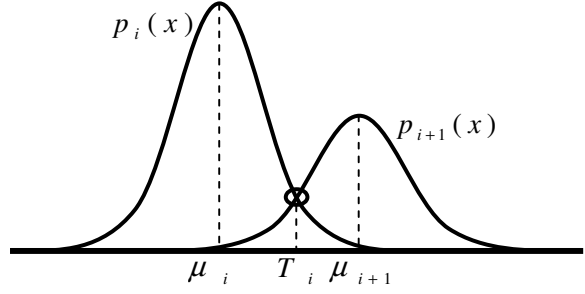

**Figure 1.** Determination of the threshold points.

Although in the literature several methods to obtain the thresholds for a gray level image have been reported as in [27], it is known that lots of these methods are computationally expensive, e.g., in the Otsu method [26] in order to find two thresholds, the number of possible combinations it is $\binom{L}{2}$. For this reason, a combination of nature inspired algorithms and classic techniques, as [26], are a good option to deal with computational cost, as were made in [31, 1, 2, 4,20, 23, 24, 25], being shown that the evolutionary-based approaches actually provide a satisfactory performance in case of image processing problems.

## 3. Differential Evolution

As mentioned before, DE algorithm was introduced by Storn and Price in 1995 [28]. Even though DE was proposed more than a decade ago, researchers' interest on this metaheuristic continues growing, due mainly to its simplicity to implement, robustness and convergence properties [28]. This algorithm is population based, and employs mutation and crossover operations; however, the most important is the mutation, which is the central procedure [2]. This operator uses the weighted difference between a randomly selected pair of parents. Also, DE uses a non-uniform crossover, which permits the individuals to share information related to promising searching areas through successful combinations information.

The first step in DE consists on initializing a uniformly distributed random population formed by a number of parents $N_p$, each one with a $D$-dimensional vector, limited by pre specified lower ($\boldsymbol{x\_low}$) and upper ($\boldsymbol{x\_high}$) limit vectors:

$$x_{i,j}^k = x\_low_j + rand()(x\_high_j - x\_low_j) \qquad (9)$$

$$j = 1, \dots, D; \quad i = 1, \dots, N_p; \quad k = 0$$

Generate a mutant vector is achieved with:

$$\boldsymbol{v}_i^k = \boldsymbol{x\_best}^k + F \cdot (\boldsymbol{x}_{r_1}^k - \boldsymbol{x}_{r_2}^k) \qquad (10)$$

$$r_1, r_2 \in \{1, 2, \ldots, N_p\}, \quad r_1 \neq r_2 \neq i$$

where $r_1$ and $r_2$, are randomly selected integer indexes, $x\_best^k$ represents the best population member found so far, and $F$ is a scaling mutation factor, usually less than 1. $N_p$ represents the number of parents. By using one or more mutant vectors and a crossover parameter, it is obtained a trial vector:

$$u_{i,j}^k = \begin{cases} v_{i,j}^k & if\ rand() \leq C_r\ or\ j = j_{rand} \\ x_{i,j}^k & otherwise \end{cases} \quad (11)$$

$$j_{rand} \in \{1, 2, \ldots, D\}, \quad 0.0 \leq C_r \leq 1.0$$

where the crossover constant delimits the use of some parts belonging to the mutant vector that will be part of the trial vector. In the last part of the algorithm it is used a selection operator, in order to improve solutions, according to:

$$x_i^{k+1} = \begin{cases} u_i^k & if\ f(u_i^k) < f(x_i^k) \\ x_i^k & otherwise \end{cases} \quad (12)$$

In this equation, $f$ represents the cost function show in equation 3; all the steps are repeated until certain criteria is reached, usually a maximum iteration number $N_{max}$, although in this work is also used a minimum distance obtained.

## 4. Particle Swarm Optimization

Particle Swarm Optimization (PSO) was created as a general purpose optimizer in 1995, when Kennedy and Eberhart joined their efforts in order to simulate the social behavior of some species; those efforts evolved until the simulation became a general purpose optimizer whose main idea lies behind the individuals in a swarm [5]. The way Kennedy and Eberhart achieved that was by modeling the synchronized behavior of swarms of birds as a function of the individual efforts to maintain optimal distances among each individual and its neighbors. The first use as an optimizer of PSO was the training of a three layer neural network for the problem of X-OR with excellent results; since then, PSO has been put to work on classification, nets communication, segmentation, design and control, among many other applications [6].

In this work we have used a modification of Clerc's PSO [9], where a constriction parameter is used. In that PSO version, the two main governing equations are:

$$v_i^{k+1} = \omega_k \cdot \left( v_i^k + c1 \cdot rand() \cdot \left( p_{best_i} - x_i^k \right) + c2 \cdot rand() \cdot \left( g\_best - x_i^k \right) \right) \quad (13)$$

$$x_i^{k+1} = v_i^{k+1} + x_i^k$$

In the first part of the algorithm, the particles' positions and velocities are randomly initialized:

$$x_{i,j}^k = x\_low_j + rand()(x\_high_j - x\_low_j) \tag{14}$$

$$v_{i,j}^k = v\_low_j + rand()(v\_high_j - v\_low_j)$$
$$j = 1,2,...,D; \quad i = 1,2,...,N_p; \quad k = 0$$

considering $rand()$ is a uniformly distributed random number, $x\_high_j$ and $v\_high_j$ are the superior limits that positions and velocities can reach; $x\_low_j$ and $v\_low_j$ are the respective inferior limits, and $N_p$ states the particle's number. The next part deals with knowing $\boldsymbol{p\_best_i}$ known as the best particle found at $i$-th position until iteration $k$, and $\boldsymbol{g\_best}$ represents the best global particle found so far, considering all the population. Once the aforementioned values are found, velocities and position of all particles are actualized by using equation 13, taking into account that $\omega_k$ is the constriction parameter as proposed in [1], modified with:

$$\omega_k = \omega_0 \cdot \exp\left(\frac{-\rho \cdot k}{N_{max}}\right) \tag{15}$$

Usually $c1 \approx c2 \approx 2$, and others values were taken the values proposed in [1] as we were using such proposal. It is important to notice that the dynamic constriction parameter is a slightly difference to the original Clerc's algorithm, in which the constriction parameter is static. In this case, $\omega_0$ is an initial constriction value, $\rho$ is a control value, $k$ is the actual iteration and $N_{max}$ represents the maximum number of iterations.

## 5. Artificial Bee Colony Optimization

As mentioned before, Karaboga [6] developed an algorithm based on the behavior of honey bees, called ABC. The ABC pseudo code is shown next:

```
1. Initialize food sources
      2. Repeat
            3. Each employed goes to a food source in its memory
                  3a. Determines a neighbor source
                  3b. Evaluates nectar
                  3c. Return to hive and dances
            4.Each onlooker watches the dance
                  3a. Chooses one of the sources, considering the intensity of
                      dance
                  3b. Goes to the food source selected
                  3a. Determines a neighbor source
```

```
            3b. Evaluates nectar
    5. The food sources abandoned are determined
            5a. Abandoned food sources are replaced by new ones discovered
                by scouts
    6. The best food source until this iteration is saved
 7. Go to step 2 until a certain criteria is reached.
```

The initial food sources are randomly initialized, by the formula:

$$x_{i,j}^k = x\_low_j + rand()(x\_high_j - x\_low_j) \qquad (16)$$

$$j = 1,2,\dots,D; \quad i = 1,2,\dots,N_{fs}; \quad k = 0$$

being considered that $x\_high_j$ and $x\_low_j$ are the upper and lower limits where the function to optimize is defined; $N_{fs}$ is the number of food sources, $D$ states for dimensions and $k$ is the actual iteration.

Next in the algorithm, each employed bee is sent to a randomly selected food source and a neighbor is determined randomly in order to produce a modification to the source stored in its memory:

$$b_{i,j} = x_{i,j} + \varphi_{i,j}(x_{i,j} - x_{l,j}) \qquad (17)$$

where $\in \{1,2,\dots,N_{eb}\}$, $j \in \{1,2,\dots,D\}$, $l$ is randomly selected, $i \neq l$ and $\varphi_{i,j}$ is a random number between $[-1,1]$. If this modification produces a food source outside the limits, then it is set to the appropriate limit ($x\_high_j$ or $x\_low_j$).

Both, the source in memory and the modified one are evaluated; the bee memorizes the new position and forgets the old one. Later, employed bees return the hive and dances; and onlooker bees will choose a food source to exploit according to a probability's function:

$$q_i = \frac{fit_i}{\sum_{j=1}^{N_{fs}} fit_j} \qquad (18)$$

where $fit_i$ represents the fitness of solution $i$, evaluated by the employed $i$, calculated by:

$$fit_i = \begin{cases} \frac{1}{f(x_i)+1} & if\ f(x_i) \geq 0 \\ 1 + abs(f(x_i)) & elsewhere \end{cases} \qquad (19)$$

Later, again a neighbor is determined by the onlooker by means of equation 17, both food sources are evaluated and the best is memorized. Finally, one scout is generated at the end of each iteration in order to explore for new food sources. The cycle is repeated both until a minima distance is reached or a maximum iterations number.

## 6. Experimental Results

In all experiments, each candidate solution holds the elements: $[P_1^i, P_2^i, P_3^i, \mu_1^i, \mu_2^i, \mu_3^i, \sigma_1^i, \sigma_2^i, \sigma_3^i]$, the population inside the three algorithms were the same size and a distance stop criterion was determined experimentally. The main idea behind using the same population size for all the experiments lays in the fact that in order to do the comparisons in similar circumstances for all the algorithms, the main parameter to consider is the function´s evaluation number. Such parameter is considered the most expensive in the computational cost. Table 1 shows the general parameters utilized by the algorithms, whereas Table 2 shows particular parameters used by each one. It is important remark that all the experiments were performed using a desktop computer with AMD Athlon II 2.9GHz microprocessor, with 4GB in RAM and programmed in Matlab 7.12.0.

| Parameter | DE, PSO, ABC: value | Observations |
|---|---|---|
| L | 256 | Number of gray levels |
| D | 9 | Dimensionality of each candidate solution |
| w | 3 | Penalty associated with $\sum_{j=1}^{K} P_j = 1$ |
| $N_p$ | 90 | Population size |
| $N_{max}$ | 200 | Maximum number of iterations |
| $x\_high$ | [0.5,0.5,0.5,L-1,L-1,L-1,(L-1)/2,(L-1)/2,(L-1)/2] | Higher limits of candidate $i$ |
| $x\_low$ | [0,0,0,0,0,0,0,0,0] | Lower limits of candidate $i$ |
| k | Variant | Actual iteration |
| K | 3 | Number of classes to find |
| T | 2 | Number of thresholds to find |

**Table 1**. General parameters used in DE, PSO and ABC.

| ABC | | DE | | PSO | | |
|---|---|---|---|---|---|---|
| Onlooker's Number | Employee's number | F | Cr | $\omega_0$ | $\rho$ | $c1, c2$ |
| $N_p/2$ | $N_p/2$ | 0.25 | 0.8 | 3 | 4.6 | 2.0 |

**Table 2**. Particular parameters used in DE, PSO and ABC.

In the first part of the experiments the main idea was to find the correct separation between two objects and the background in real images, as well as the time spent by each algorithm taking in account the number of function evaluations; considering that, images related with blood cells were utilized in order to find the best approximation, and in which three (Figure 2) classes are visually found.Such image database was taken from [17], and the experiments were repeated 1000 times for each of 295 smear blood images taken from ALL-IDB database; we only consider such images and not all the 367 because the three classes must be more or less well defined in order to get good results with the proposal, as it can be seen in Figure 2.

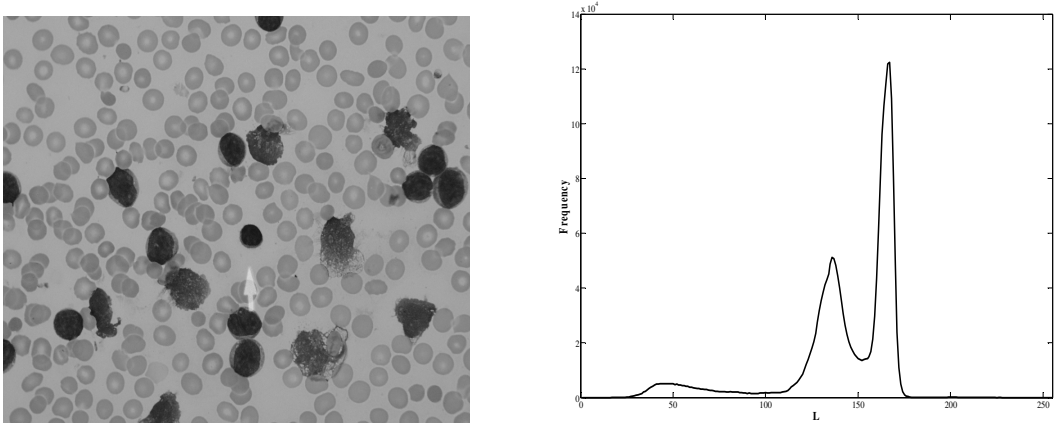

**Figure 2**. Example of experimental imageand its associated histogram.

The cells' nucleus, in darker gray, represents an object class, whereas the two remaining classes are the less dark cells, or red blood cells, and the background. After running the three algorithms, some situations can be drawn in a statistical fashion, as can be seen in Table 3; for instance, ABC and DE are more consistent in finding the minimum Hellinger distance, although DE shows a slightly better performance than ABC in such sense. In third place is PSO, with a bigger standard deviation than DE and ABC and with a mean distance also bigger. In the experiments, an iteration is considered completed when all the individuals are evaluated and modified according to the particular rules of each algorithm. To the types of problems considered in this paper, the best behavior related to iterations number belongs to DE with a mean of 76, followed by ABC, PSO and Otsu. As can be seen, evaluations of objective functions and execution times are direct consequence of the number of iterations and therefore the performance of the algorithms is the same as the aforementioned. It is also important remark that Otsu method is taken as reference in evaluation of the objective function, and the best algorithm evaluates almost 90% less than such method and therefore, meta-heuristic techniques are computationally cheaper than Otsu.

Even though several artifacts can be appreciated in the segmented images, the results were good enough to consider objects well separated from the background, based in the fact that this segmentation was done taking into account only one feature: the pixel's gray level. However, this feature is not enough to qualify results as good; considering that issue, the next experimental part is related with a quality measure based on a Hausdorff distance. Results are shown in Table 4 and discussed below it.

| Technique | Hellinger distance | | Iterations | | Objective function evaluations | | Execution time(S) | |
|---|---|---|---|---|---|---|---|---|
| | μ | σ | μ | σ | μ | σ | μ | σ |
| ABC | 0.1161 | 0.0023 | 95 | 23 | 8586 | 2090 | 0.7819 | 0.1871 |
| DE | 0.1161 | 0.0022 | 76 | 17 | 6954 | 1595 | 0.4980 | 0.1146 |
| PSO | 0.1534 | 0.0950 | 107 | 151 | 9556 | 13656 | 0.5328 | 0.7283 |
| OTSU | NA | NA | NA | NA | 65535 | NA | 6.0937 | 0.0219 |

**Table 3.** Optimization statistic results of ABC, DE, PSO and Otsu over ALL-IDB database.

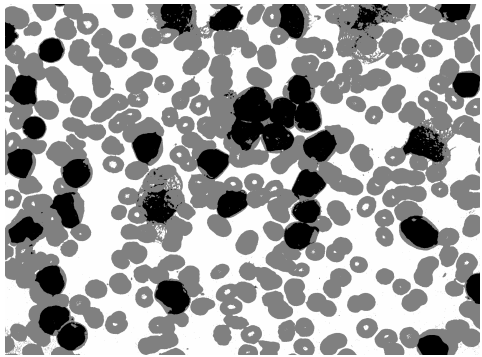
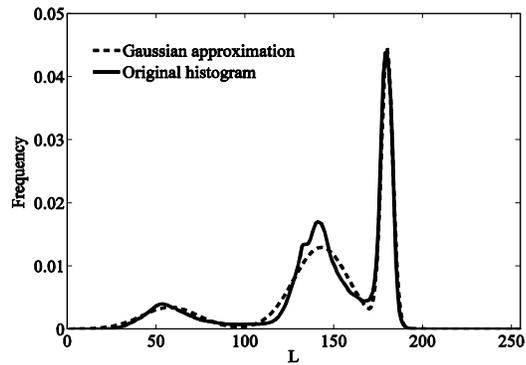

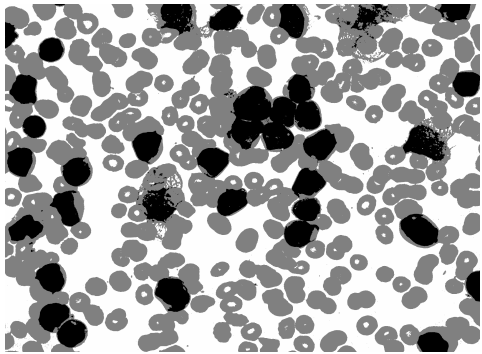
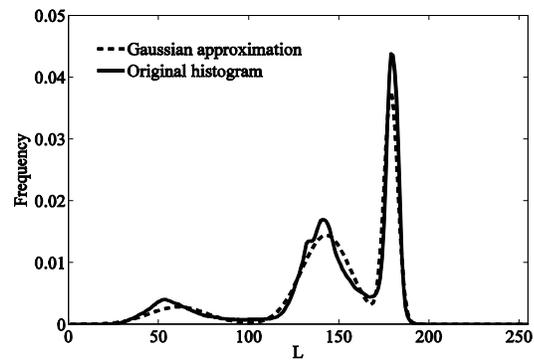

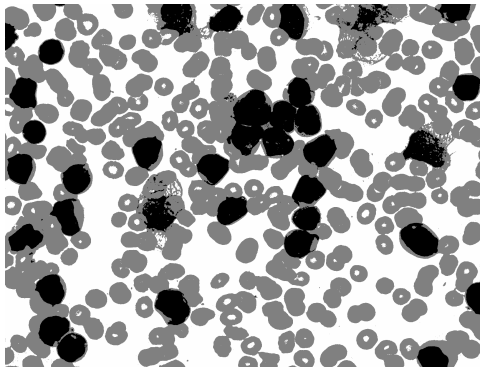
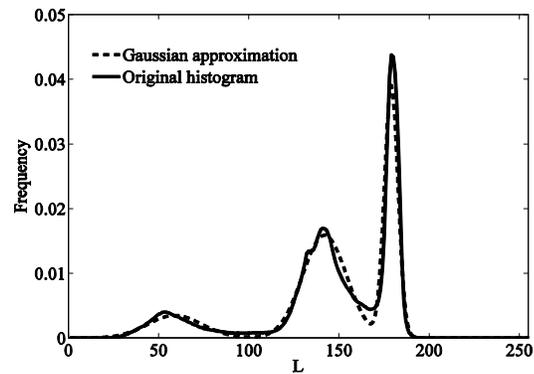

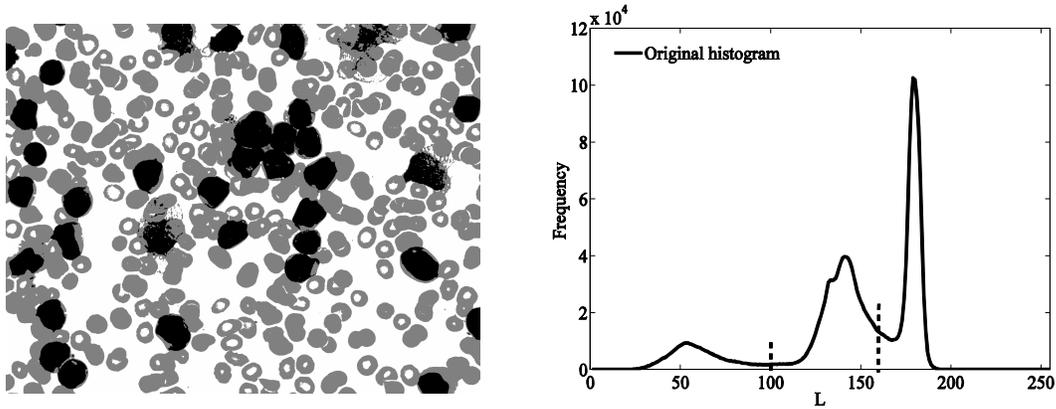

**Figure 3**. Results of the segmentation process corresponding, from top to bottom, to ABC, DE, PSO and Otsu.

| Technique | Hausdorff distance | |
|---|---|---|
| | μ | σ |
| ABC | 2.0600 | 0.0654 |
| DE | 2.1064 | 0.4577 |
| PSO | 2.1947 | 0.2365 |
| OTSU | 2.4655 | 2.6779e-015 |

**Table 4.** Quality statistic results of ABC, DE, PSO and Otsu over ALL-IDB database.

The images used in the second part of the experiments are composed of ALL-IDB images together their correspondent manually segmented ground truths images, as examples shownin Figure 4.

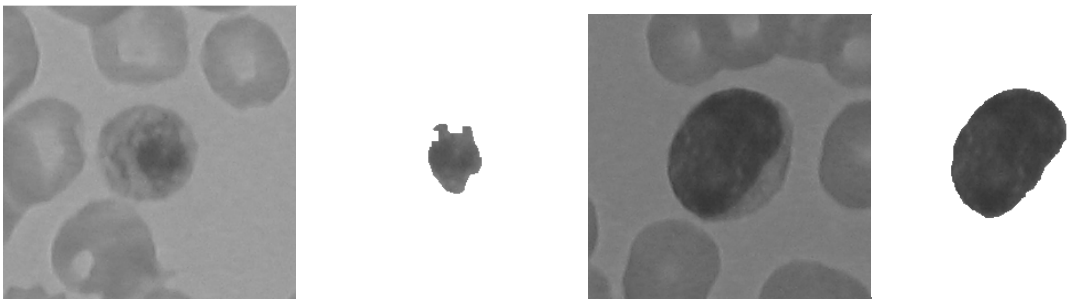

**Figure 4**. Examples of images used in the second experimental part.

By taking only the mean values of obtained Hausdorff distance, it can be seen that ABC has the best performance, followed by DE, PSO and Otsu. In such a sense, although DE outperforms the rest of the meta-heuristics in the optimization process, in quality terms ABC has the best results in the segmentation part, due mainly to mixture of Gaussian functions obtained by each algorithm are not exactly the same even though the Hellinger distance could be the same.

## 7. Conclusions

This paper presentsanempirical comparison of three state of the art nature inspired algorithms to perform image thresholding by a mixture of Gaussian functions. It is assumed that the intensity distributions for objects and background within the image obey Gaussian distributions with different means and standard deviations.

Thestatistical analysis of the results shows a superior performance of DE not only in minimizing the Hellinger distance between the original and the candidate histogram but also performing such a minimization in less evaluations of the mentioned cost function based on distance, although the best quality performance of the meta-heuristic algorithms belongs to ABC, followed by DE, PSO and Otsu. Even though Otsu is an exhaustive technique, in quality terms has not the best performance in segmentation terms. In the experiments, were used the database of smear blood images proposed in [17].

Future work includes comparisons by using other distance measures with the best meta-heuristic -the considered computationally cheaper-, found in this work, in order to determine if the convergence of the algorithm is improved according the cost function used.